\documentclass[sigconf]{acmart}

\usepackage{bm}

\AtBeginDocument{%
  \providecommand\BibTeX{{%
    \normalfont B\kern-0.5em{\scshape i\kern-0.25em b}\kern-0.8em\TeX}}}

\acmConference[KDD '20]{KDD '20: ACM Special Interest Group (SIG) on Knowledge Discovery and Data Mining Conference Workshops}{August 23--27, 2020}{San Diego, CA}
\acmBooktitle{KDD '20: ACM Special Interest Group (SIG) on Knowledge Discovery and Data Mining Conference Workshops,
  August 23--27, 2020, San Diego, CA}

\begin{document}

\title[Resilient In-Season Crop Type Classification in Multispectral Satellite Observations]{Resilient In-Season Crop Type Classification in Multispectral Satellite Observations using Growth Stage Normalization}


\author{Hannah Kerner}
\email{hkerner@umd.edu}
\orcid{0000-0002-3259-7759}
\affiliation{%
  \institution{University of Maryland, College Park}
}
\author{Ritvik Sahajpal}
\affiliation{%
  \institution{University of Maryland, College Park}
}
\author{Sergii Skakun}
\affiliation{%
  \institution{University of Maryland, College Park}
}
\author{Inbal Becker-Reshef}
\affiliation{%
  \institution{University of Maryland, College Park}
}
\author{Brian Barker}
\affiliation{%
  \institution{University of Maryland, College Park}
}
\author{Mehdi Hosseini}
\affiliation{%
  \institution{University of Maryland, College Park}
}
\author{Estefania Puricelli}
\affiliation{%
  \institution{University of Maryland, College Park}
}
\author{Patrick Gray}
\affiliation{%
  \institution{Duke University}
}

\renewcommand{\shortauthors}{Kerner, et al.}

\begin{abstract}
Crop type classification using satellite observations is an important tool for providing insights about planted area and enabling estimates of crop condition and yield, especially within the growing season when uncertainties around these quantities are highest. As the climate changes and extreme weather events become more frequent, these methods must be resilient to changes in domain shifts that may occur, for example, due to shifts in planting timelines. In this work, we present an approach for within-season crop type classification using moderate spatial resolution (30 m) satellite data that addresses domain shift related to planting timelines by normalizing inputs by crop growth stage. We use a neural network leveraging both convolutional and recurrent layers to predict if a pixel contains corn, soybeans, or another crop or land cover type. We evaluated this method for the 2019 growing season in the midwestern US, during which planting was delayed by as much as 1-2 months due to extreme weather that caused record flooding. We show that our approach using growth stage-normalized time series outperforms fixed-date time series, and achieves overall classification accuracy of 85.4\% prior to harvest (September-November) and 82.8\% by mid-season (July-September). 
\end{abstract}




\maketitle

\section{Introduction}
\label{sec:intro}
Knowledge of crop-specific planted area, conditions, and expected yields is critical for informing agricultural markets and decision making as well as ensuring food security globally. Official reports by agricultural ministries, such as the USDA’s Crop Progress and Condition Reports, are widely followed by market analysts and traders to inform decision-making around agricultural markets and futures. Thus, uncertainty in public information including official reports for major food-producing countries results in uncertainty and volatility in agricultural markets and prices \cite{Lehecka2014}. This uncertainty can arise when they are shocks to the system, e.g., record flooding or the COVID-19 pandemic in the 2019 and 2020 growing seasons respectively. Market tensions and perceptions of increased uncertainty about agricultural production caused by policy decisions and reduced access to information influence market prices and decisions that can disrupt market stability and food security.

Earth observations (EO) data provide a valuable complementary source of information about global crop progress and conditions, enabling timely analysis at field scales throughout the growing season to supplement traditional data collection and reporting efforts. Crop type masks that provide classifications for crop-specific land cover are critical inputs for agricultural monitoring quantities including estimates of acreage, production, condition, and yield. Prior work on crop type classification has achieved promising results using time series information to differentiate spectrally-similar crop types. However, models trained to distinguish crop types based on temporal patterns may not generalize well to inputs that have shifted in the time domain, for example due to delayed growth or planting as a result of extreme weather events or climate change. 

An example of this is the 2019 growing season in the midwestern United States. In spring 2019, heavy snowfall and rains caused unprecedented flooding across the midwestern states and forced farmers to plant crops 1-2 months later than normal \cite{Irwin2019}. While such extreme weather events are historically rare, they are becoming more frequent and exacerbate uncertainty and confusion for markets and decision-makers. Thus, for crop type classification models to be used operationally, it is important that they 1) be resilient to shifting growing seasons, and 2) provide predictions within the growing season when uncertainty around planting and production is highest.

In this paper, we present a method for crop type classification that uses time series inputs at key growth stages detected at the pixel level to mitigate the effects of domain shift caused by shifts in planting timelines. Additionally, this method enables in-season predictions of crop type at each detected growth stage. We demonstrated the performance of this method for mapping corn and soybean crops in a study region of the midwestern US in northern Illinois, and showed that our growth stage-normalized approach gives better classification accuracy than using fixed time series inputs. While this study is limited to crop type classification, our approach could be used for distinguishing other land cover types that exhibit unique phenological signatures, such as other plant or tree species, where time series classification approaches may be affected by seasonal shifts or climate change.

\section{Related work}
\label{sec:related_work}
Most prior work on crop type classification incorporates spectral and temporal information in the input to a machine learning classifier. Random forests and decision trees are widely used for crop type classification due to their good performance in many case studies and their interpretability (e.g., \cite{Hariharan2018,Neetu2019,Palchowdhuri2018,Saini2018,Shelestov2019,Shukla2018,Song2017,Sonobe2017_Landsat,Sonobe2017_Sentinel,Sonobe2018,Viskovic2019,MingWang2019,Wang2019}), though the input features extracted from the satellite observations vary. For example, Song et al. \cite{Song2017} used bagged decision trees to classify time series statistics extracted from optical Landsat spectral bands, whereas Wang et al. \cite{Wang2019} used the coefficients of harmonic regression as input features for a random forest classifier. Kernel methods including support vector machines (SVMs) and kernel-based extreme learning machines \cite{Pal2013} have also shown good performance for crop type classification (e.g., \cite{Feng2019,Neetu2019,Saini2018,Sonobe2017_Landsat,Sonobe2017_Sentinel,Wagstaff2006,MingWang2019})---for example, Feng et al. \cite{Feng2019} extracted more than 100 features including spectral, textural, and phenological properties of optical observations for SVM classification. 

One limitation of traditional machine learning methods such as random forests and SVMs is that they require input features to be extracted from the original data to achieve good performance, which may be sub-optimal for discriminating between target classes and often requires domain expertise. To address these limitations, recent work (e.g., \cite{Brandt2019,Cai2018,Chakrabarti2019,Garnot2019,Ji2018,Kussul2017,Russwurm2017,Zhong2019}) has demonstrated improved classification performance using neural network-based approaches which automatically learn useful features for discriminating between target classes using a standard optimization procedure \cite{LeCun2015}. Zhong et al. \cite{Zhong2019} showed that a convolutional neural network (CNN) with 1D convolutions outperformed a long short term memory (LSTM) network \cite{Hochreiter1997} for classifying time series inputs of only one spectral index (enhanced vegetation index, or EVI). Cai et al. \cite{Cai2018} showed that a deep neural network had better performance when information from multiple spectral bands and indices were included in the time series inputs. Kussul et al. \cite{Kussul2017} demonstrated better performance using an ensemble of 2D CNNs compared to 1D CNNs using time series from multiple optical bands and synthetic aperture radar (SAR) observations. Ru{\ss}wurm et al. \cite{Russwurm2017} showed that an LSTM gave better performance than a 2D-CNN for multispectral time series inputs that also included a neighborhood of pixels around the predicted pixel. Similarly, Garnot et al. \cite{Garnot2019} found that temporal features extracted in recurrent layers were more important than spatial features extracted in 2D convolutional layers for parcel-level crop type classification, but that hybrid methods combining recurrent and convolutional layers had the best performance. Brandt \cite{Brandt2019} showed that a network with capsule and attention layers in addition to CNN and LSTM layers performed better for crop type classification than a 2D CNN, LSTM, and combined CNN-LSTM. Following the success of 3D CNNs for modeling spatio-temporal relationships in videos, Ji et al. \cite{Ji2018} showed that 3D CNNs sometimes improved performance over other methods including SVMs and 2D CNNs. Together, these studies suggest that better crop type classification performance can be achieved with deep learning methods that combine both recurrent and convolutional layers to model patterns in the spectral, spatial, and temporal dimensions of remote sensing observations.

Many prior crop type classification studies have not assessed model generalization to future years that may exhibit growing season timelines that differ from the years used in training. Additionally, most prior studies rely on observations spanning the complete growing season, and thus do not provide within-season predictions of crop types. These limitations are barriers to operational adoption of crop type classification techniques. In this study, we present an approach for within-season crop type classification that is robust to shifts in growing season timelines by dynamically selecting input observations based on crop phenology. To assess the performance of this approach in future years that exhibit domain shift with respect to the training data, we evaluated our method for the 2019 growing season in the midwestern US in which planting was delayed by as much as 1-2 months. 

\section{Study area and reference data}
\label{sec:dataset}

We focused our study on a region in northern Illinois that covers approximately 12,056 km$^2$ including Knox and Stark counties as well as most of Peoria, Henry, and Bureau counties. We chose this region because northern Illinois was one of the regions most affected by extreme weather and floods in early 2019, and Illinois is the second-largest producer of corn in the United States \cite{USDA_Production}. Corn and soybeans are the dominant crops grown in this region, while forests, grassland, and urban development cover most area not covered by corn and soybean fields.
\begin{figure}[t]
\begin{center}
   \includegraphics[width=\linewidth]{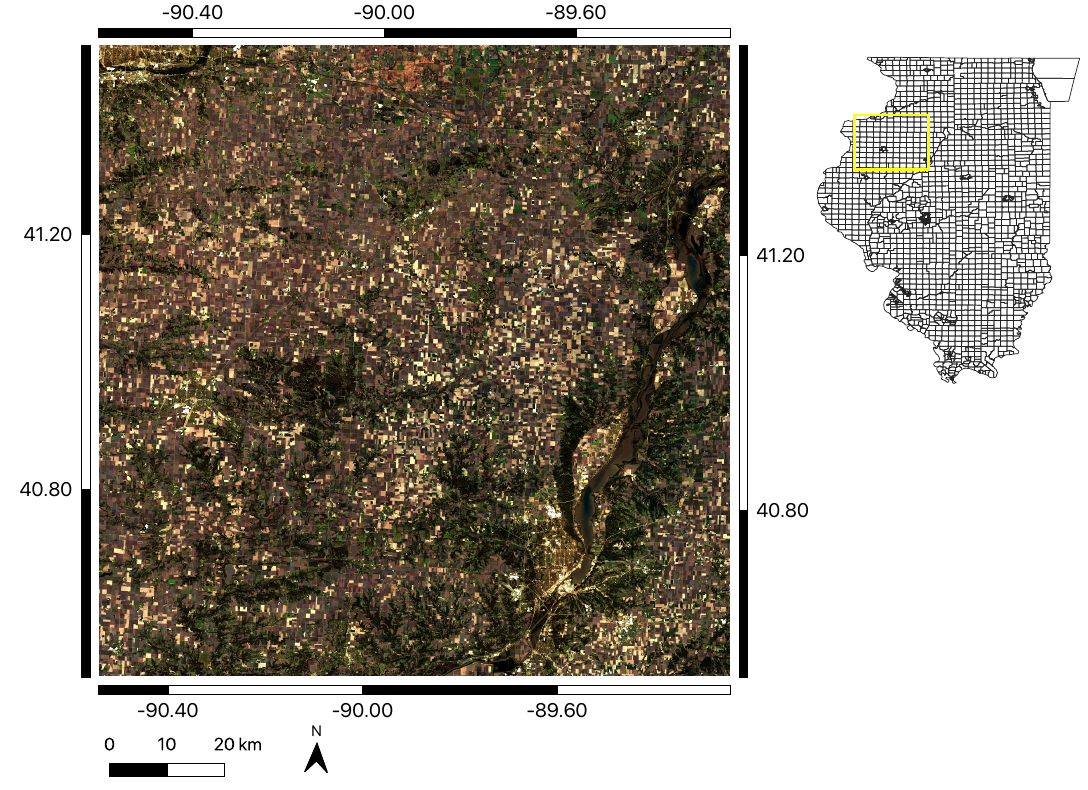}
\end{center}
   \caption{Inset: study area in northern Illinois covering 12,056 km$^2$ including Knox, Stark, Peoria, Henry, and Bureau counties (tile 16TBL in HLS reference system).}
\label{fig:study_area}
\end{figure}

We used satellite observations from the Harmonized Landsat-8 and Sentinel-2 (HLS) and Sentinel-1 datasets. HLS adjusts Sentinel-2 surface reflectance products to have the same bandpasses and spatial resolution (30 m/pixel) as Landsat-8. This allows the Landsat-8 and Sentinel-2 satellites to be treated as a virtual constellation with 30 m/pixel spatial resolution and 2-3 days revisit time \cite{Claverie2018}. Sentinel-1 acquires C-band synthetic aperture radar (SAR) observations every 6 days at the equator. We used observations acquired between 2017-2019 since the second Sentinel-2 satellite was not launched until 2017. For Sentinel-1, we used VV and VH polarisation bands (speckle-filtered and normalized by incidence angle). For HLS, we used the blue, green, red, near infrared (NIR), and short-wave infrared (SWIR) bands (Landsat-8 bands B01-B06 and Sentinel-2 bands B02, 03, 04, 08A, 11, and 12) as well as two spectral indices: normalized difference water index (NDWI) and land surface water index (LSWI). We computed NDWI and LSWI using the following equations:
\begin{equation}
    \text{NDWI} = \frac{\text{GREEN}-\text{SWIR1}}{\text{GREEN}+\text{SWIR1}}
\end{equation}
\begin{equation}
    \text{LSWI} = \frac{\text{NIR}-\text{SWIR1}}{\text{NIR}+\text{SWIR1}}
\end{equation}
We used NDWI and LSWI instead of more commonly used indices---e.g., normalized difference vegetation index (NDVI) or green chlorophyll vegetation index (GCVI)---because water content inside the leaf has been shown to an important factor for discriminating between corn and soybeans \cite{Cai2018,Feng2019}, which is captured by the SWIR bands included in the calculations for NDWI and LSWI. In our experiments we found that classification performance decreased when NDVI and GCVI were included in the input. We used the quality assessment (QA) layer provided in the HLS dataset to remove pixels with clouds and cloud shadows (bit numbers 0-3). The HLS tiles also contain no-data pixels (NaNs) in regions not covered by the satellite track for a particular date. To fill in no-data pixels and mitigate outliers in the time series data, we smoothed the time series for each pixel and each band using the Savitzky-Golay method \cite{Chen2004}.

We used the Cropland Data Layer (CDL) produced by the United States Department of Agriculture (USDA) National Agricultural Statistics Service (NASS) \cite{Boryan2011} as a surrogate for ground truth data, as has been done in prior studies (e.g., \cite{Chakrabarti2019,Wang2019}). According to the CDL, approximately 60\% of our study area is covered by corn or soybean fields, and most fields follow a corn-soybean rotation. Labels sourced from the CDL cannot be considered ground truth since they are predicted using a combination of ground truth labels and a decision tree model. However, the CDL typically has high accuracy for corn and soybeans in midwestern states \cite{Boryan2013}, with user's and producer's accuracies between 90-98\% for corn and soybeans in Illinois estimated for CDL 2017-2019 \cite{CDL_Metadata}. Lark et al. \cite{Lark2017} recommends post-processing steps to reduce the number of errors in the CDL labels, including aggregating classes and filtering out within-field speckle and other heterogeneity. We aggregated all classes except corn and soybeans into an ``other'' class. We applied a $3\times3$ homogeneity filter to identify non-homogeneous regions (caused by within-field speckle) to be masked out during training. 

\begin{figure*}
  \includegraphics[width=\textwidth]{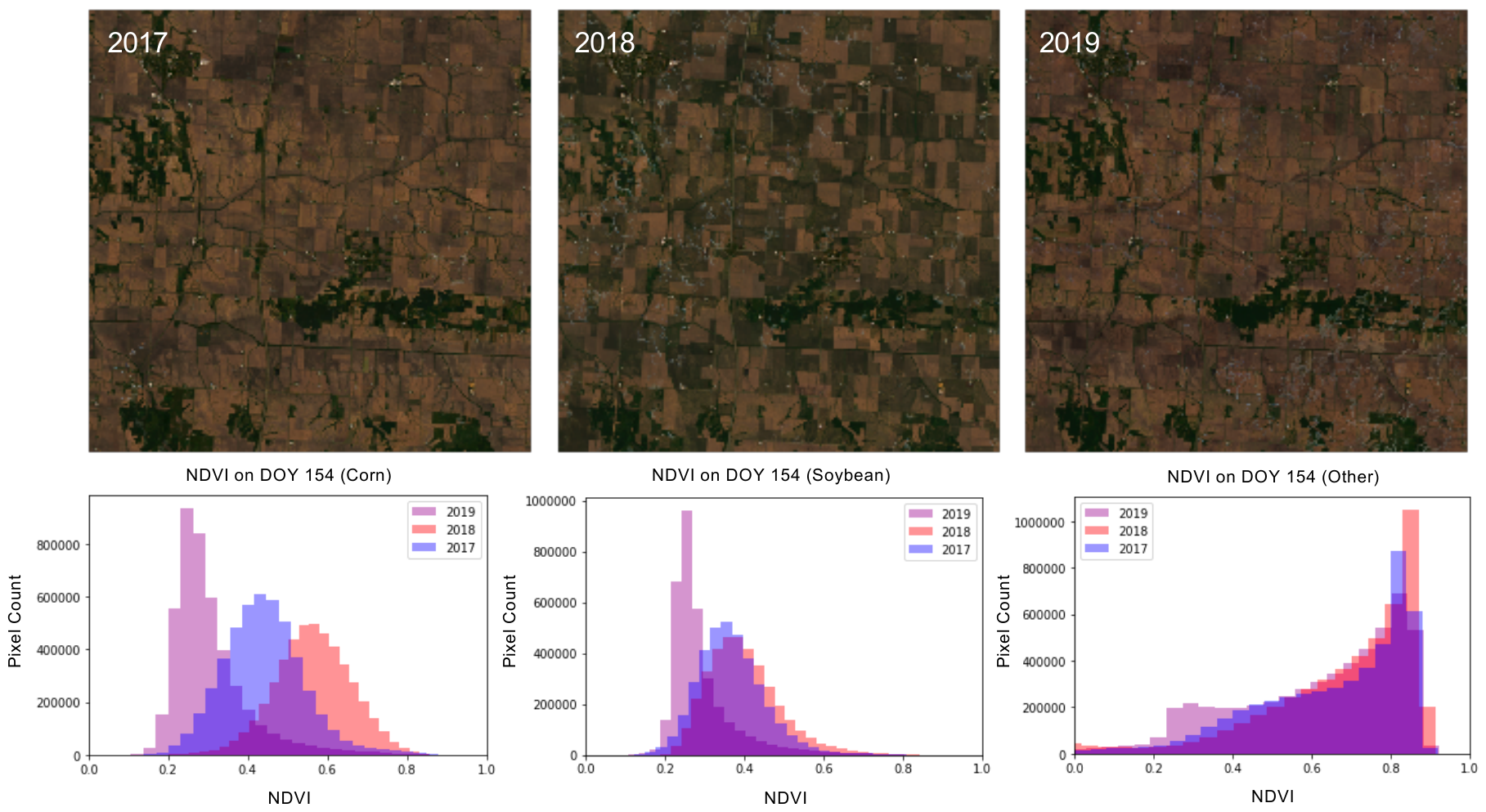}
  \caption{True color Harmonized Landsat and Sentinel-2 (HLS) image on DOY 154 (June 3) in 2017, 2018, and 2019 over study area subset (inset in Figure \ref{fig:maps}) (top). Different levels of vegetation at the same time of year indicate different planting timelines. Histograms of NDVI for pixels classified as corn, soybean, or other in the CDL show the distribution in 2019 is shifted significantly lower than 2017-2018 for corn and soybeans, but mostly consistent for other.}
  \label{fig:domain_shift}
\end{figure*}

\section{Approach}
\label{sec:approach}
Time series observations of crops from satellite images may have different temporal patterns from year to year depending on the planting timeline followed each year, which may change due to extreme weather or other climate factors. While methods that use the time series from the full growing season have been shown to have good performance for discriminating crop types, these methods will likely have poor performance for seasons in which planting timelines are different than in the training seasons (i.e., domain shift). Furthermore, most methods do not enable predictions of crop types within the growing season, when uncertainty around crop production is most uncertain. We propose a new approach for crop type classification that addresses domain shift by normalizing inputs by crop growth stage. In each pixel, we automatically select observations at key growth stages rather than using observations from the same dates for all pixels. These key observations are then fed to a deep neural network that combines recurrent and convolutional layers to capture spectral, spatial, and temporal relationships in the input observations.

\subsection{Growth Stage Normalization}
While specific growth stages vary between crop types, all crops generally contain three key growth stages: greenup, during which crops emerge from the ground and undergo early vegetation; peak growth, during which crops have reached their maximum growth and leaf area index (LAI), and begin to ripen; and senescence, during which crops are drying down before they are harvested. For many crops including corn and soybeans, these growth stages are apparent in the time series of vegetation indices like NDVI (e.g., \cite{Becker2019,Wardlow2006}), though regional planting timelines may vary from year to year due to weather or other climate-related factors that influence soil conditions or temperature \cite{Hatfield2015}. For this reason, it is important that observations are compared at the same growth stage, which may not be at the same time \cite{Skakun2019}. Observations of the same crop on the same day but in different years will not be comparable if the planting timelines each year were different, resulting in domain shift. Figure \ref{fig:domain_shift} shows the distribution of NDVI values in corn, soybean, and other pixels as well as true color images on the same day of year (DOY) in 2017, 2018, and 2019. In 2019, the distribution of NDVI values at the same DOY is significantly lower for both corn and soybeans due to planting delays caused by extreme flooding, while the other class distribution showed only minor differences (likely due to minor crops that were also delayed). 

\begin{figure}[t]
\begin{center}
   \includegraphics[width=0.8\linewidth]{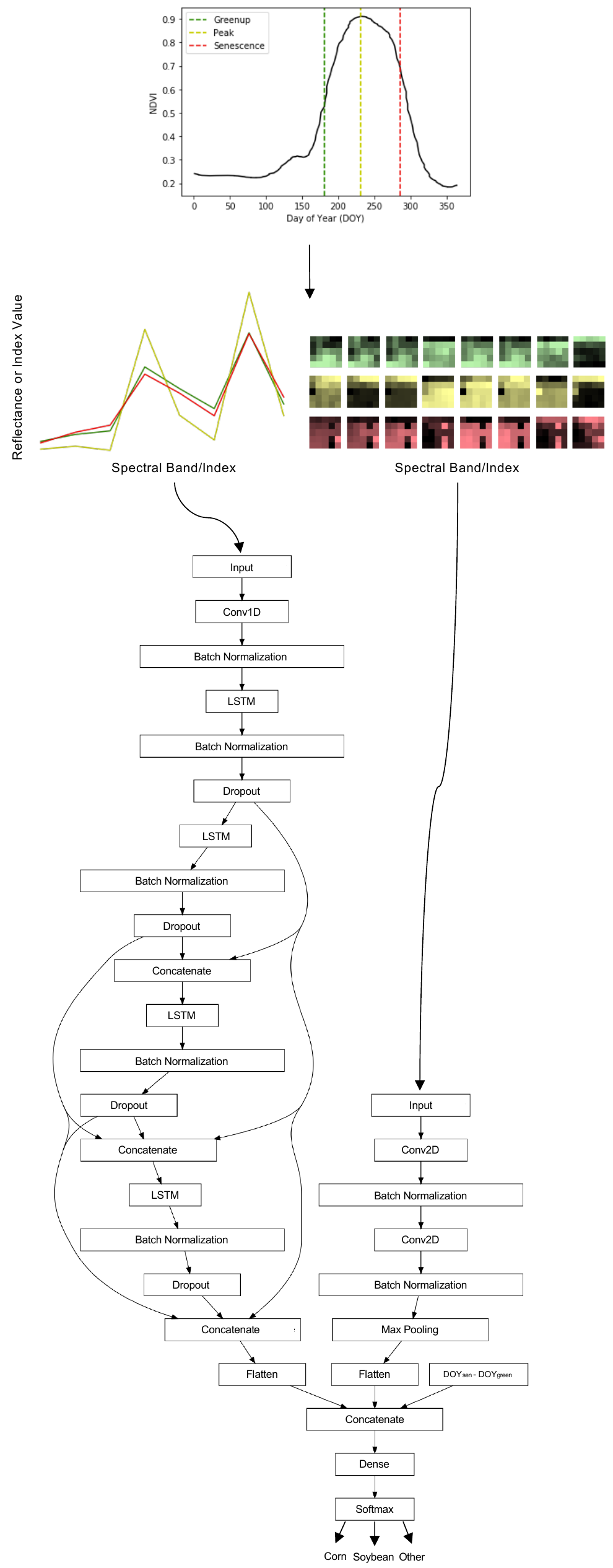}
\end{center}
   \caption{Crop type classification approach. Spectral and spatial information from DOYs selected based on key growth stages (greenup, peak growth, and senescence) are input to neural network with CNN and LSTM branches.}
\label{fig:architecture}
\end{figure}

Since observations at the same growth stage should be comparable regardless of planting timeline, our approach is to distinguish crop types using satellite observations acquired at key growth stages rather than using fixed dates or the full growing season. For each pixel in our dataset, we computed the approximate date of greenup, peak growth, and senescence using the NDVI time series. We defined the greenup DOY to be the day after which the slope of the NDVI time series curve is highest (i.e., increasing fastest) within the first half of the time series. We defined the DOY of senescence to be the day (following the greenup DOY) after which the slope of the NDVI time series curve is the lowest (i.e., the most negative, thus decreasing fastest). We defined the DOY of peak growth to be the day at which NDVI reaches its maximum value between the greenup and senescence DOYs. Figure \ref{fig:architecture} shows an example NDVI time series with greenup, peak growth, and senescence DOYs.

\subsection{CNN-LSTM Classifier}
Prior studies detailed in Section \ref{sec:related_work} suggest that neural network architectures that use both convolutional and recurrent layers to capture spatial and temporal patterns in data (respectively), in addition to using observations from multiple spectral bands, should have the best performance for crop type classification compared to alternate methods. Our model architecture (shown in Figure \ref{fig:architecture}) consists of two branches: a 2D CNN branch to capture spatial and/or textural patterns, and an LSTM branch to capture temporal patterns. The CNN and LSTM branches are joined by a dense layer before the final softmax prediction layer. Since the number of growing days is a distinguishing factor between corn and soybeans, we included a third input to the dense layer joining the CNN and LSTM branches that is the scalar number of days elapsed between the senescence DOY and greenup DOY, i.e.: $\delta=\text{DOY}_{\text{senescence}}-\text{DOY}_{\text{greenup}}$. 

\textbf{Recurrent branch.} The first layer in the recurrent branch is a 1D convolution layer (8 filters, $1\times3$ kernels) followed by four LSTM layers with 16 units in all layers. All LSTM layers are densely connected, meaning the output of all previous LSTM layers is included in the input in addition to the output of the prior layer. Each LSTM layer is followed by a batch normalization \cite{Ioffe2015} layer and dropout \cite{Srivastava2014} layer (randomly dropping 30\% of units). We used the rectified linear unit (ReLU) activation function for all layers. The input to the LSTM branch consists of 6 HLS spectral reflectance values, 2 HLS spectral index values, and 2 SAR polarisation values (see Section \ref{sec:dataset} and Figure \ref{fig:architecture}) for each pixel acquired on the greenup, peak, and senescence DOYs---i.e., $\bm{x}_i^{\text{LSTM}} \in \mathbb{R}^{M\times D}$ where $M=10$ is the number of channels and $D=3$ is the number of timesteps.

\textbf{Convolutional branch.} The convolutional branch consists of two 2D convolution layers, each having 64 filters and $3\times3$ kernels with 1-pixel strides. We found that fewer or more than two convolutional layers did not improve performance for the validation dataset. We used ReLU activation for both layers, and each layer is followed by a batch normalization layer (dropout did not improve model generalization in this branch). We included a $2\times2$ max pooling layer after the two convolution layers. The input to the CNN branch consists of $5\times5$ image patches in which the central pixel is the pixel to be classified for all 10 channels and 3 timesteps. We also experimented with larger patch sizes (odd dimensions between $7\times7$-$13\times13$), but found this did not improve validation accuracy. The multispectral patches for each timestep are concatenated to produce a 3D input tensor: $\bm{x}_i^{\text{CNN}} \in \mathbb{R}^{k\times k\times MD}$ where $k=5$ is the size of the input patch, $M=10$ is the number of channels, and $D=3$ is the number of timesteps.

\textbf{Dense layer.} The output tensors of the CNN and LSTM branches are flattened and joined by a dense (fully-connected) layer with 64 units. This dense layer is followed by the final softmax output layer that gives posterior probabilities of inputs belonging to the \textit{corn}, \textit{soybean}, or \textit{other} output classes.

We implemented our CNN-LSTM architecture using the Keras deep learning library with the TensorFlow backend. The code for our model and experiments (detailed in Section \ref{sec:experiments}) is available at \url{https://github.com/nasaharvest/croptype-mapping-gsn}.

\begin{table*}
\caption{Performance metrics (overall accuracy (OA), per-class precision/user's accuracy (UA) and recall/producer's accuracy (PA)) for each crop type classification method using input observations from fixed and growth stage-normalized (GS-Norm) dates. \textbf{Bold} indicates best performance across all methods.}
\begin{center}
\begin{tabular}{|l|c|c|c|c|c|c|c|c|c|c|}
\hline
\multicolumn{1}{|c|}{Measure} & \multicolumn{1}{|c|}{CNN-LSTM$_\delta$} & \multicolumn{2}{|c|}{CNN-LSTM} & \multicolumn{2}{|c|}{CNN} & \multicolumn{2}{|c|}{LSTM} & \multicolumn{2}{|c|}{Random Forest}\\
 & GS-Norm & GS-Norm & Fixed & GS-Norm & Fixed & GS-Norm & Fixed & GS-Norm & Fixed \\
\hline\hline

OA & \textbf{85.4} & 82.8 & 55.9 & 82.1 & 65.1 & 76.3 & 48.8 & 83.7 & 82.8 \\
UA/Precision & \text{ } & \text{ } & \text{ } & \text{ } & \text{ } & \text{ } & \text{ } & \text{ } & \text{ } \\
\hspace{3mm}\textit{Corn} & 91.1 & \textbf{91.2} & 28.4 & 89.1 & 55.8 & 87.6 & 38.9 & 86.2 & 81.1 \\
\hspace{3mm}\textit{Soybean} & 83.2 & 86.7 & 34.6 & \textbf{91.2} & 44.8 & 86.8 & 90.8 & 89.2 & 82.1 \\
\hspace{3mm}\textit{Other} & 83.1 & 76.7 & 83.9 & 75.0 & 85.9 & 68.2 & \textbf{89.1} & 79.9 & 84.6 \\
PA/Recall & \text{ } & \text{ } & \text{ } & \text{ } & \text{ } & \text{ } & \text{ } & \text{ } & \text{ } \\
\hspace{3mm}\textit{Corn} & 79.2 & 72.6 & 10.7 & 77.6 & 39.7 & 60.2 & \textbf{95.6} & 81.5 & 84.4 \\
\hspace{3mm}\textit{Soybean} & \textbf{81.0} & 77.7 & 57.8 & 67.6 & 58.2 & 68.2 & 2.6 & 70.2 & 74.7 \\
\hspace{3mm}\textit{Other} & 93.1 & 94.0 & 90.5 & 94.7 & 89.7 & \textbf{94.1} & 40.3 & 93.7 & 86.5 \\

\hline
\end{tabular}
\end{center}

\label{tab:e1_perf}
\end{table*}

\begin{figure*}[t]
\begin{center}
   \includegraphics[width=\linewidth]{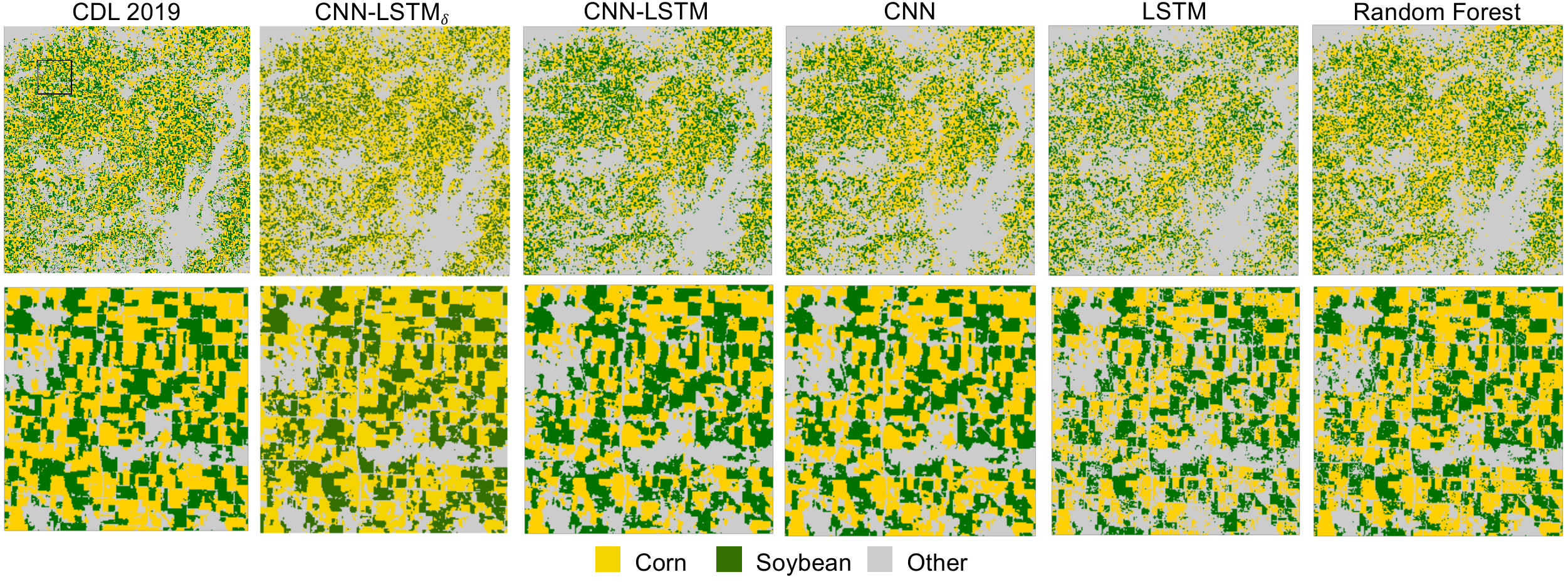}
\end{center}
   \caption{First column: Cropland Data Layer (CDL) for 2019 showing reference labels of \textit{corn}, \textit{soybean}, and \textit{other}. Remaining columns: predicted crop types for all four late-season classification models. Top row shows the full study area and bottom row shows subset of the full study area (outlined in black in CDL 2019 image) to show finer-scale differences between predicted maps and CDL.}
\label{fig:maps}
\end{figure*}

\section{Experiments}
\label{sec:experiments}

Our goal for all experiments in this study is to assess model generalization to observations acquired in future years that were not part of the training dataset, in order to assess the resilience of each method to time domain shifts. We used observations from 2017 and 2018 in the training and validation datasets and from 2019 in the test dataset, since 2017 and 2018 follow typical planting timelines for corn and soybeans in the US midwest while 2019 saw delayed planting (discussed in Section \ref{sec:intro}). We partitioned our study region (shown in Figure \ref{fig:study_area}) into quadrants and used pixels in the northwest, northeast, and southeast quadrants for training and the southwest quadrant for validation. In all experiments, we used the validation dataset to assess when learning plateaued and ensure the model did not overfit to the training dataset. We conducted three experiments to evaluate the performance for crop type classification in the beginning, middle, and end (before harvest) of the growing season. 

\begin{table*}
\caption{Performance metrics (overall accuracy, per-class precision/user's accuracy (UA) and recall/producer's accuracy (PA)) for each crop type classification method for early-season and mid-season classification experiments. \textbf{Bold} indicates best performance across mid-season classification and early-season classification methods.}
\begin{center}
\begin{tabular}{|l|c|c|c|c|c|c|c|}
\hline
\multicolumn{1}{|c|}{Measure} & \multicolumn{1}{|c|}{CNN-LSTM$_\delta$} & \multicolumn{1}{|c|}{CNN-LSTM} & \multicolumn{2}{|c|}{CNN} & \multicolumn{1}{|c|}{LSTM} & \multicolumn{2}{|c|}{Random Forest}\\
 & Mid-Season & Mid-Season & Early-Season & Mid-Season & Mid-Season & Early-Season & Mid-Season \\
\hline\hline

Overall accuracy & 80.0 & 72.8 & 55.5 & \textbf{82.8} & 72.7 & \textbf{69.0} & 76.5 \\
Precision (UA) & \text{ } & \text{ } & \text{ } & \text{ }  & \text{ } & \text{ }  & \text{ }  \\
\hspace{3mm}\textit{Corn}  & \textbf{94.9} & 93.2 & \textbf{80.5} & 89.3 & 84.2 & 62.7 & 73.2 \\
\hspace{3mm}\textit{Soybean} & 82.6 & \textbf{91.2} & 51.0 & 83.6 & 89.8 & \textbf{63.1} & 87.6 \\
\hspace{3mm}\textit{Other} & 72.9 & 62.3 & 55.6 & \textbf{78.7} & 64.1 & \textbf{77.0} & 75.3 \\
Recall (PA) & \text{ } & \text{ } & \text{ } & \text{ } & \text{ } & \text{ } & \text{ }\\
\hspace{3mm}\textit{Corn}  & 63.3 & 54.8 & 10.5 & \textbf{75.0} & 53.7 & \textbf{75.6} & 76.0 \\
\hspace{3mm}\textit{Soybean}  & 75.7 & 55.9 & \textbf{49.7} & \textbf{77.8} & 60.0 & 41.1 & 52.5 \\
\hspace{3mm}\textit{Other}  & 95.8 & \textbf{97.5} & \textbf{94.6} & 92.0 & 95.5 & 81.1 & 91.7 \\

\hline
\end{tabular}
\end{center}

\label{tab:e2_perf}
\end{table*}

\subsection{Late-season classification}
In the late-season classification experiment, classifier inputs contain observations acquired at all three growth stages: greenup, peak growth, and senescence. The mean senescence DOY for 2017-2019 is $278 \pm 28$. This is around the time that crops are harvested in this region (September-November) and production estimates remain uncertain, thus our model predictions would enable estimates several months before the CDL is made available publicly by the USDA (typically February the following year). To assess the contribution of each component in our model, we compared the performance of our model (denoted CNN-LSTM$_\delta$) with the CNN branch only, the LSTM branch only, and the CNN-LSTM without the scalar $\delta$ input. To assess the benefit of normalizing inputs by growth stage rather than using fixed DOYs, we evaluated model performance using inputs from fixed DOYs (except CNN-LSTM$_\delta$ since $_\delta$ would be constant across all inputs). For the first date we used DOY 153, 149, and 152 for 2017, 2018, and 2019 respectively since greenup typically occurs around late May/early June; for the second date we used DOY 210, 209, and 211 since peak growth is typically reached around late July; and for the third date we used 274, 271, and 274 since senescence typically occurs around late September. Since random forests are widely used for crop type classification in prior work \cite{Orynbaikyzy2019}, we also compared our method to a random forest with 900 estimators (chosen based on highest accuracy from 5-fold cross validation for estimators between 100-1000 with increments of 100). The input to the random forest included the 6 optical bands, 2 SAR bands, 2 spectral index values, and $_\delta$. We implemented the random forest using Scikit-learn \cite{scikit-learn} in python. 

We trained all models for this and subsequent experiments for 25 epochs (after which validation accuracy plateaued) using mini-batch sizes of 4096, cross entropy loss, and the Adam optimizer with default settings \cite{Kingma2014}. To select training samples, we randomly sampled 520,000 pixels from the available training pixels (i.e., those not masked out by the homogeneity filter) and 10,000 from the available validation pixels. To mitigate class imbalance, we subsampled majority classes to have the same number of samples as the minority class (96,459 training and 5,517 validation samples per class). We used the same random sample and seed for all experiments. We observed empirically that validation accuracy decreased when a substantially larger or smaller number of training examples was used. We report the performance metrics (overall accuracy as well as per-class user's accuracy/precision and producer's accuracy/recall) for the five models using both growth-stage normalized inputs and fixed DOYs in Table \ref{tab:e1_perf}. We used the Sieve tool in QGIS to reduce within-field speckle in the CDL labels before computing performance metrics. In Figure \ref{fig:maps}, we illustrate the qualitative performance in the predicted classification maps for our study area compared to the CDL reference labels.  

Table \ref{tab:e1_perf} shows that the growth-stage normalized CNN-LSTM$_\delta$ model has the highest overall accuracy for the 2019 test data. Table \ref{tab:e1_perf} also shows that for all models, growth-stage normalized inputs give better performance than fixed DOYs. The CNN-LSTM architecture consists of two branches---an LSTM and 2D-CNN branch---with outputs joined by a dense (fully-connected) layer. The only difference between this and CNN-LSTM$_\delta$ is the addition of a third input to this dense layer, which is the scalar difference between the senescence and greenup DOYs, intended to capture the number of growing days. Table \ref{tab:e1_perf} shows that including $\delta$ increases accuracy by 2.6\%. The performance of the CNN branch-only model performs much better (by 5.8\%) than the LSTM branch-only model, which suggests that spatial information is more informative for discriminating crop types than temporal information in this experiment. While prior work found recurrent layers to have greater influence on performance than convolutional layers (e.g., \cite{Brandt2019,Garnot2019}), the difference is likely due to the shorter time series (3 timesteps) used in this study. The random forest had lower overall accuracy than the CNN-LSTM$_\delta$, but higher overall accuracy than the other neural network models. This is likely because the random forest includes the scalar $\delta$ input while the CNN-LSTM, CNN, and LSTM models do not, but does demonstrate that a random forest can achieve comparable performance to deep learning methods for this problem. Table \ref{tab:e1_perf} also shows that the difference between the random forest with growth-stage normalized and fixed DOY inputs is much lower than for the neural network models. This suggests that random forests may extrapolate better to changes in planting timelines than neural network approaches, since the random forest does not explicitly model temporal or spatial patterns that rely on the ordering of inputs. 

\subsection{Early and mid-season classification}
Since all models in the late-season classification experiments described in the previous section performed better with growth stage-normalized inputs, we did not report results using fixed DOYs in the two remaining experiments for early-season and mid-season classification. We used the same experimental setup as for the late-season classification, but only included greenup and peak growth stage inputs in the mid-season classification and only greenup for the early-season classification experiments. For the CNN-LSTM$_\delta$ model, we computed $\delta$ as the difference between greenup and peak DOYs instead of greenup and senescence as in the late-season model. We only evaluated the CNN model and the random forest for early-season classification since there is no temporal information to be captured by recurrent layers. Table \ref{tab:e2_perf} gives the performance metrics for each model in the mid-season and early-season classification experiments. While the random forest results in Table \ref{tab:e1_perf} were achieved including $\delta$ in the model input, we found that the random forest had significantly higher accuracy for the mid-season classification experiment when $\delta$ was not included in the input (32.8\% with $\delta$, 76.5\% without). Thus the random forest results in Table \ref{tab:e2_perf} were achieved without the $\delta$ input.

As expected, the overall accuracy for all models decreased when the senescence DOY was excluded from the input in the mid-season classification experiment. This drop in accuracy is primarily due to low accuracy for the corn and soybean classes, as we observed that the recall for the other class remained high regardless of the number of timesteps. The CNN had the highest overall accuracy (82.8\%) for mid-season classification, emphasizing that recurrent layers are less effective given fewer temporal observations. The mean peak growth DOY in our experiments is $216 \pm 27$, so this classification performance could be achieved between July-September within the growing season. 

For the early-season classification experiment, the overall accuracy is significantly lower when only the greenup timestep is included in the input to the CNN (27.3\% lower than in mid-season). This underscores the conclusion from prior studies and this study that crop type classification performance improves significantly when observations from multiple timesteps are used (even if only a few dates). The random forest accuracy is also lower in early-season than mid-season (7.5\% lower), but this difference is much lower than for the CNN. We also found that with only one timestep, the random forest performs better than the CNN. Thus, while the neural network approaches are more effective when additional growth stages are included later in the season, the random forest has the best performance given limited data in the beginning of the season. This could also be a result of the complexity of the neural network models, which is more suitable for the higher-dimension inputs in the later-season experiments. The mean greenup DOY in our experiments is $144 \pm 37$, so early-season predictions could be made using the random forest with 69.0\% accuracy between April-June near the start of the growing season.

\section{Discussion}

Most prior work on crop type classification has focused on post-season classification and have not addressed the effect of domain shift on model performance in future years not included in the training dataset. To be used operationally, models must be resilient to domain shifts that may occur in future seasons---e.g., due to delayed planting---as the global climate changes and extreme weather events become more frequent. Operational methods also can't rely on collecting more data each year to capture these shifts in training examples, e.g., due to the COVID pandemic or in countries where security is compromised. Rather than using observations from a fixed time period or set of dates, our approach normalizes inputs to be captured at the same key crop growth to ensure comparability between observations acquired in different years regardless of the planting timeline. Our experiments in Section \ref{sec:experiments} showed the best within-season classification accuracy of 85.4\% was achieved using our proposed CNN-LSTM$_\delta$ model with growth-stage normalized inputs late in the season prior to harvest (September-November), and that overall accuracy of 82.8\% could be achieved in mid-season (July-September) and 69\% at the beginning of the growing season (April-June) using the growth-stage normalized CNN and random forest respectively. In future work we will determine if this remains true when the models are applied to larger, more geographically diverse areas in the US Corn Belt states.

Our experiments confirmed prior findings that neural network architectures that used both recurrent and convolutional layers with multispectral, multi-temporal observations gave the best crop type classification performance (e.g., \cite{Brandt2019,Garnot2019}). Our late-season classification model included inputs from only three timesteps; in future work we will explore using more timesteps between the greenup and senescence DOYs to make better use of the recurrent layers. Although one of the main benefits of neural networks is that useful features can be automatically learned from data without the need for domain-specific feature engineering, we found that introducing some domain-specific information in the form of growth stage-normalized inputs enabled all models (including random forests) to be more robust to patterns in future (test) seasons not captured in the training data. Consistent with prior studies (e.g., \cite{Hariharan2018,Kenduiywo2017,Kussul2017,Orynbaikyzy2019,Shelestov2019,Sonobe2017_Sentinel}), we also found that including SAR inputs from Sentinel-1 in addition to optical inputs significantly improved classification accuracy; in future work, we will explore additional SAR parameters in addition to VV and VH polarization that can be extracted from SAR data. Finally, we plan to investigate aggregating pixel-level observations or predictions at the field-level, which has been shown to improve classification performance in prior work (e.g., \cite{Cai2018,Chakrabarti2019}). 

\section{Conclusion}

Crop type classification is an important technique for providing information about planted area and enabling estimates of crop condition and forecasted yield, especially within the growing season when uncertainties around these quantities are highest. As the global climate changes and extreme weather events become more frequent, crop type classification methods (for both within-season and post-season classification) must be resilient to domain shifts (e.g., due to changes in planting timelines) that can adversely affect predictive performance. Models that provide reliable predictions without the need to collect new training data each season is especially important when ground surveys are difficult or impossible, e.g., due to conflict-related insecurity or travel restrictions imposed from the COVID-19 pandemic.

We proposed an approach for within-season crop type classification that uses remote sensing inputs normalized by crop growth stage to ensure compatibility across seasons even when domain shifts may occur. We used a neural network leveraging both convolutional and recurrent layers to predict whether pixels belong to the \textit{corn}, \textit{soybean}, or \textit{other} class. We evaluated this method for the 2019 growing season in the midwestern United States, during which planting was delayed by as much as 1-2 months due to extreme weather that caused record flooding. We showed the highest classification accuracy of 85.4\% could be achieved near harvest between September-November, 82.8\% accuracy mid-season between July-September, and 69.9\% between May-July in the beginning of the growing season. Our study was limited to a region spanning several counties in northern Illinois, and our next step is to scale our approach to all of the US Corn Belt states. We will assess operational use of this approach during the 2020 growing season to provide information to help mitigate uncertainty in agricultural markets caused by the COVID-19 pandemic. 

\bibliographystyle{ACM-Reference-Format}
\bibliography{references}

\end{document}